\documentclass{article}

\usepackage[preprint]{neurips_2025}

\usepackage[utf8]{inputenc} 
\usepackage[T1]{fontenc}    
\usepackage{hyperref}       
\usepackage{url}            
\usepackage{booktabs}       
\usepackage{amsfonts}       
\usepackage{nicefrac}       
\usepackage{microtype}      
\usepackage{xcolor}         

\usepackage{comment}
\usepackage{bm}
\usepackage{multirow}
\usepackage{makecell}
\usepackage{textcomp}
\usepackage{graphicx}
\usepackage{longtable}
\usepackage{subcaption}
\usepackage{lipsum}
\usepackage{multicol}
\usepackage{amsmath}
\usepackage{booktabs}
\newcommand{\greencheck}{\color{green}\checkmark}
\usepackage{pifont}
\newcommand{\redxmark}{\color{red}\text{\sffamily X}}  
\usepackage{algorithm}
\usepackage{algpseudocode}

\usepackage{wrapfig}

\title{Lossless Token Sequence Compression via Meta-Tokens}

\author{
  John Harvill$^\spadesuit$, Ziwei Fan$^\spadesuit$, Hao Wang$^\spadesuit$$^\diamondsuit$, \textbf{Luke Huan}$^\spadesuit$\textbf{,} \\ \textbf{Anoop Deoras}$^\spadesuit$\textbf{,} \textbf{Yizhou Sun}$^\clubsuit$$\dagger$, \textbf{Hao Ding}$^\spadesuit$ \\
  $^\spadesuit$AWS AI Labs, $^\clubsuit$Amazon, \\
  $^\diamondsuit$Rutgers University, $\dagger$University of California Los Angeles \\
  \texttt{harvnhoj@amazon.com} \\
}

\begin{document}

\maketitle

\begin{abstract}

Existing work on prompt compression for Large Language Models (LLM) focuses on lossy methods that try to maximize the retention of semantic information that is relevant to downstream tasks while significantly reducing the sequence length. In this paper, we introduce a task-agnostic lossless compression technique similar to LZ77 that makes it possible to reduce the input token sequence length on average by
27\% and 18\% for the two evaluation tasks explored here. Given that we use transformer-based LLMs, this equates to 47\% and 33\% less encoding computation, respectively, due to the quadratic nature of attention. The token sequence transformation is trivial to reverse and highlights that no semantic information is lost in the process. We evaluate our proposed approach on two tasks that require strict preservation of semantics/syntax and demonstrate that existing lossy compression methods perform poorly in this setting. We find that our lossless compression technique produces only a small gap in performance compared to using the uncompressed input and posit that larger models and an expanded computing budget would likely erase the gap entirely.

\end{abstract}

\section{Introduction}
\label{introduction}

The past several years have led to an adoption of generative language models at an unprecedented scale. As these models grow in terms of parameters and dataset size, new abilities continue to be unlocked. Large Language Models (LLM) are becoming better reasoners \cite{wei2022chain, shao2024deepseekmathpushinglimitsmathematical, zhang2024llm, guan2025deliberativealignmentreasoningenables, pan2024training} and are changing how people write \cite{levine2024students}, code \cite{tian2023chatgpt}, and search \cite{spatharioti2023comparing}. Given that LLMs will likely play an ever-increasing role in our lives, the cost of token usage and inference requirements become critical factors for widespread use. As models scale, so does computational cost.
For example, the recent paradigm shift of transferring training compute to inference compute has led to significant performance gains on problem-solving tasks \cite{snell2024scaling} at the expense of latency and token usage through additional LLM calls needed to search the solution space.

One way to combat increased cost/latency is prompt compression, where the goal is to maintain necessary information for a downstream task while simultaneously reducing input sequence length. There has been a recent surge in research on prompt compression that covers a variety of different modeling conditions. Some approaches seek to optimize prompts under the condition that the LLM's parameters are not accessible \cite{jiang2023llmlingua, jiang2023longllmlingua, pan2024llmlingua, chuang-etal-2024-learning}. Others focus on the gains that can be made via architectural changes to the underlying LLM and continuing to fine-tune using compression \cite{mu2024learning, chevalier-etal-2023-adapting, ge2023context}.

While previous works have been able to compress input sequences to a small fraction of their original length \cite{jiang2023llmlingua, pan2024llmlingua}, the downstream tasks explored in these works inherently allow for significant prompt compression due to the sparsity of information required to solve each task. For example, in passage-based Question Answering (QA), the answer to the question is usually contained within a small span of text, making the other information unnecessary. As we will demonstrate, for tasks where every part of the input contains important information, any amount of lossy compression can lead to the inability of an LLM to solve the task.
\begin{wrapfigure}{r}{0.52\textwidth}
    \centering
    \includegraphics[width=0.51\textwidth]{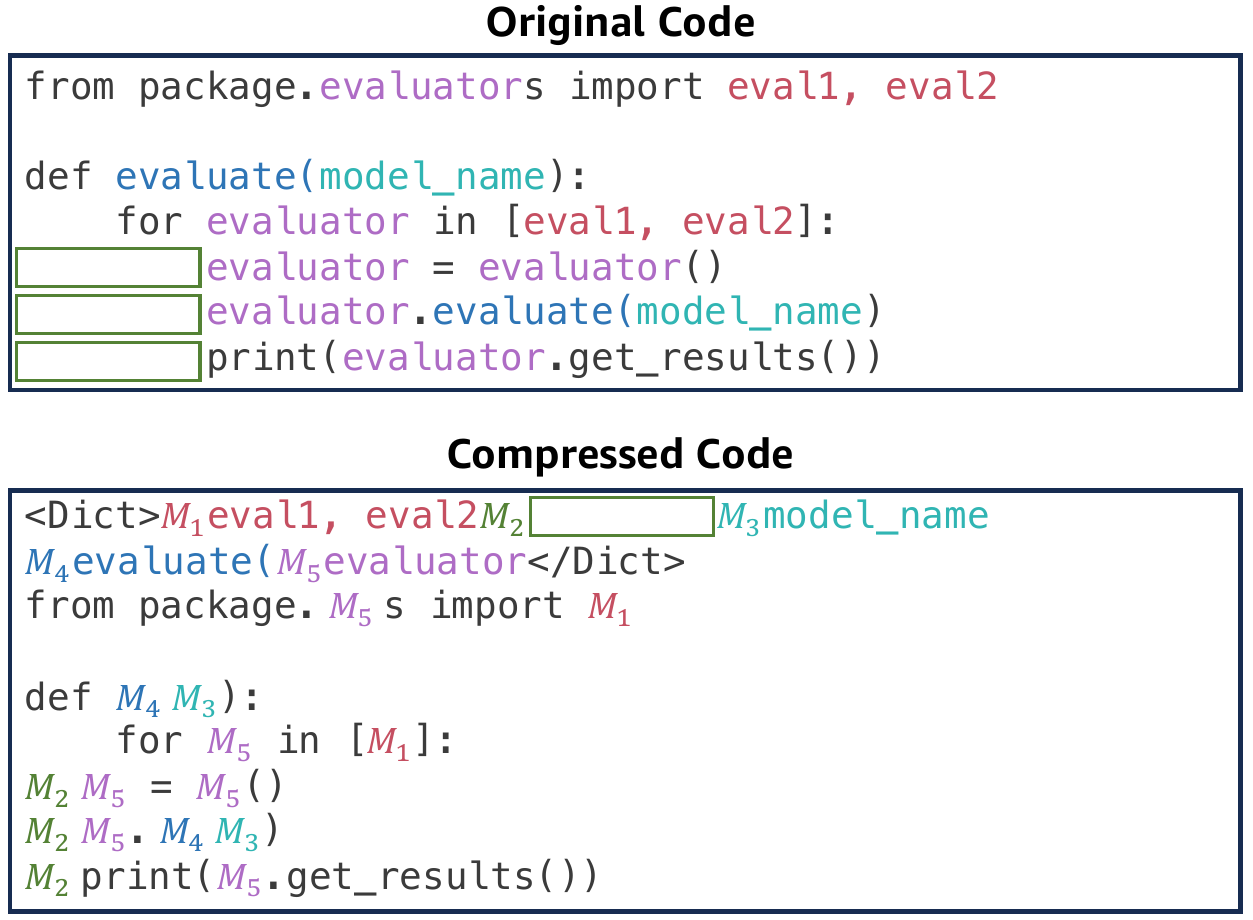}
    \caption{Example of swapping repeated subsequences with meta-tokens in a Python code snippet. Each unique meta-token is paired with a unique subsequence to form the dictionary, which is prepended to the compressed input. The green box represents spacing.
    }
    \vspace{-5pt}
    \label{fig:fig1}
\end{wrapfigure}
We explore two tasks to demonstrate this idea, and there are numerous practical scenarios in which preservation of all text characters is necessary for proper sequence understanding. These include essay/email editing, code translation, Code Reasoning Understanding and eXecution (CRUX) \cite{gu2024cruxevalbenchmarkcodereasoning}, code repair, and unit test generation. Our approach to compression is lossless and task-agnostic, making it attractive for general use cases.

Our main contribution is the proposal of \textit{\textbf{L}ossless \textbf{T}oken \textbf{S}equence \textbf{C}ompression} (\texttt{LTSC}). The key idea behind \texttt{LTSC} is that there are many repetitions of multi-token subsequences in long prompts which makes the representation of any given subsequence inefficient (see Figure \ref{fig:fig1}). By replacing each instance of a repeated subsequence with a single meta-token, we can reduce the total sequence length. To indicate the mapping of the meta-token to the original subsequence, we can construct a dictionary of meta-token/subsequence pairs and prepend the dictionary to the compressed prompt. We use the term ``meta-token'' to indicate that these tokens function as \textbf{placeholders} and can \textit{dynamically represent different token subsequences} for each prompt. We will demonstrate that fine-tuning on tasks with this compression format teaches the LLM to understand how meta-tokens replace instances of their corresponding subsequence in the original sequence, which allows token sequences to be compressed without any loss of information. Additionally, our proposed approach is flexible in that the choice of whether to apply compression can be made at inference time, and the only model architecture change needed is adding a relatively small number of vocabulary tokens (meta-tokens).
\vspace{-3pt}

\section{Related Work}
\label{related_work}

There exists a large body of work on prompt compression, which is covered extensively in the survey paper from \citet{li2024promptcompressionlargelanguage}. Prompt compression can be used to lower costs by reducing
\begin{wrapfigure}{r}{0.52\textwidth}
    \centering
    \includegraphics[width=0.51\textwidth]{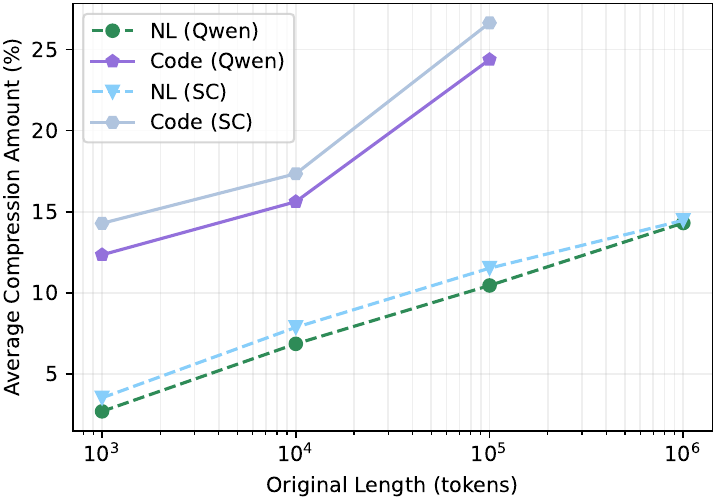}
    \caption{Compression amounts by sequence length for Natural Language (NL) and code domains with two tokenizers (Qwen: Qwen2.5-Coder, SC: StarCoder2).}
    \vspace{-5pt}
    \label{fig:comp_amount}
\end{wrapfigure}
token usage, improve generation latency, or make information denser under a fixed context length budget. Some existing work allows control over how much the prompt is compressed \cite{jiang2023llmlingua, pan2024llmlingua}, but for our proposed approach, the maximum amount of compression for a given prompt depends on the prompt itself. In this paper, we compare existing work to ours when maximally compressing every prompt and leave optimization under different use cases (fixed budget, latency, etc.) to future work. We provide a cursory overview of prompt compression in this section, with a more exhaustive discussion left to Appendix \ref{related_work_appendix}. We note that, to the best of our knowledge, we are the first to propose lossless compression for prompting. Throughout the paper, we will use the terms "compression rate" and "compression amount", where amount $A$ refers to the fraction by which the sequence length is reduced, i.e. $A = 1-|T^{'}|/|T|$ where $T^{'}$ is the compressed version of token sequence $T$,\footnote{The sequence format we propose includes an instance-specific dictionary of compression pairs and the compressed sequence. Throughout the paper, when discussing the amount of sequence length reduction, the dictionary length is included in the calculation of the compressed sequence length.} and compression rate $R = 1 - A$.

\noindent\textbf{Hard Prompt Compression} methods are limited to using tokens solely from the original LLM vocabulary \cite{li-etal-2023-compressing, zhang2024adacompextractivecontextcompression, jung-kim-ieee, shandilya2024tacorltaskawareprompt, liskavets2024promptcompressioncontextawaresentence, liu-etal-2023-tcra}. This results in two forms of prompt compression: \textbf{1)} Dropping uninformative tokens. \textbf{2)} Generating a shorter token sequence with similar semantics to the original. The first form is popular and makes the most sense from a latency perspective. Given an existing sequence, deciding which tokens to drop becomes a binary classification problem and most computation for each token's classification can be performed in parallel. Generating a summarized version of the prompt introduces significant latency overhead but can lead to superior performance. Our proposed approach introduces little latency (see Figure \ref{fig:runtime}) while also perfectly preserving semantics, making it an improvement over both existing forms of hard prompt compression.

\noindent\textbf{Soft Prompt Compression} involves the tuning of new vocabulary tokens or existing model parameters to capitalize on a more efficient representation of the prompt \cite{mu2024learning, chevalier-etal-2023-adapting, ge2023context, li2024500xcompressorgeneralizedpromptcompression, gao2024unifyingdemonstrationselectioncompression}. These approaches draw inspiration from prompt and prefix tuning \cite{lester-etal-2021-power, li-liang-2021-prefix}, where the goal is to fine-tune a small number of parameters that allow a frozen LLM to perform a new task without being exposed to that task during pre-training. The main benefits of \texttt{LTSC} are the fact that compression is lossless and fine-tuning is extremely simple. Unlike existing work that requires customized attention masks \cite{mu2024learning} or back-propagation through time \cite{chevalier-etal-2023-adapting}, our approach only requires additional vocabulary tokens and can easily be plugged into existing fine-tuning pipelines.

\section{Compressibility}

We first motivate our approach by showing how much compression can be achieved for different types of sequences when using \texttt{LTSC}. The compression amounts produced by two tokenizers across natural language and code domains is given in Figure \ref{fig:comp_amount}, where the amount is the percentage by which the compressed sequence is shorter than the original sequence \textit{when including the dictionary}. For natural language, we concatenate articles from Wikipedia to achieve various token sequence lengths, up to 1M tokens. For code, we use Python repositories from the RepoBench dataset \cite{liu2023repobench}, varying in length from 1k-100k tokens.

\noindent\textbf{Domain Differences}: One of the key differences to notice is the large gap between compression amounts between the natural language and code domains for token sequences of the same length. We highlight the differences between these two domains to show that the amount of compression that can be achieved is domain-specific, and can vary widely. We hypothesize that code is more compressible than natural language for several reasons: 

\noindent\textbf{1)} \textit{Syntax}: Code contains repeated syntax structures that do not have their own tokens due to the limited vocabulary size. For example, python repositories contain large amounts of indentation, common patterns such as "):\textbackslash n\textbackslash t" for function definitions or \texttt{for}/\texttt{while} loops, etc. Common code readability standards, such as putting each argument on its own line or adhering to docstring formatting guidelines, introduce additional opportunity for compression.

\noindent\textbf{2)} \textit{Repository-specific code}: Repetitive uses of repo-specific code can add redundancy to an input sequence. Structures such as import prefixes "\texttt{import package.}", function/variable names, etc. can be represented by a single meta-token which can reduce the token sequence length by a large margin.

\noindent\textbf{Tokenizer Effect}: The size of the tokenizer vocabulary also plays a role in the compressibility of a given token sequence. LLMs with smaller vocabulary sizes must represent some text sequences with multiple tokens that can be represented with one token by an LLM with a larger vocabulary. This lack of efficiency from smaller vocabularies leads to more opportunity for compression. We directly see this phenomenon when comparing the compression amounts of \texttt{StarCoder2} and \texttt{Qwen2.5-Coder} for both natural language and code domains. Given that \texttt{StarCoder2}'s vocabulary size ($\sim$50k) is much smaller than \texttt{Qwen2.5-Coder}'s ($\sim$150k), \texttt{StarCoder2}'s representation of the same text sequences will be less efficient by creating more repeated token subsequences, leading to larger amounts of compression when applying \texttt{LTSC}.

\begin{figure}
    \centering
    \includegraphics[width=0.93\linewidth]{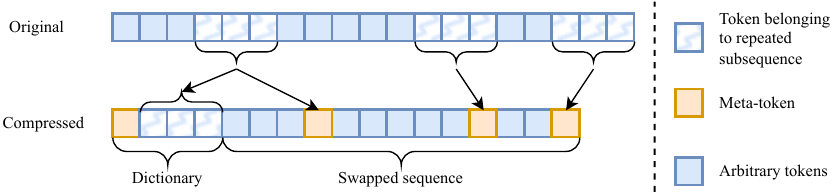}
    \caption{Visualization of the compression process. We exclude the XML tags <Dict> and </Dict> for illustration purposes since these are negligible for long sequences, and only include one subsequence/meta-token pair for simplicity. Note that the compressed sequence is shorter and that the repeated subsequence representation is reduced from $3\times3=9$ tokens in the original to $(1+3)+3=7$ tokens in the compressed version (see Equation \ref{eq:swap_condition}). The compressed representation is lossless, because we can reconstruct the original sequence by replacing the meta-tokens with their respective subsequences and removing the dictionary.}
    \label{fig:swapping}
\end{figure}

\section{Algorithm}

While tokenization compresses text significantly compared to a character based representation, tokenized sequences can still contain repetitions of multi-token subsequences. Such redundancy is inefficient, so we propose \textit{\textbf{L}ossless \textbf{T}oken \textbf{S}equence \textbf{C}ompression} (\texttt{LTSC}), a simple algorithm that replaces repetitive subsequences with single, unique meta-tokens (See Figures \ref{fig:fig1} and \ref{fig:swapping}). Our proposed algorithm is similar to the classical LZ77 compression algorithm \cite{LZ77}, where the original subsequence is referred to in the future when that subsequence appears again. Instead of using match distance and match length to represent the repetition, we use unique meta-tokens. To indicate the mapping between the original subsequence and its meta-token, we prepend a dictionary of subsquence/meta-token pairs to the beginning of the token sequence. We will show that under the right conditions, the tokens needed to represent the dictionary and the compressed sequence can be markedly less than the original number of tokens.

\noindent\textbf{Compressibility Conditions}: If a given token subsequence $T_{sub}$ of length $N$ appears $K$ non-overlapping times in the token sequence $T$, then it uses $N \cdot K$ tokens for its representation (see Figure \ref{fig:swapping}). If instead we create a meta-token $M$ to represent $T_{sub}$, we need $1 + N$ tokens to indicate the mapping of $M$ to $T_{sub}$ in the dictionary, but only $K$ instances of $M$ to represent $T_{sub}$ in the original sequence for a total of $1 + N + K$ tokens. Thus, for subsequences that satisfy the following inequality:
\begin{equation}\label{eq:swap_condition}
    N \cdot K > 1 + N + K
\end{equation}
replacing them with a meta-token results in a shorter token sequence. By solving Equation \ref{eq:swap_condition} for all integer values of $N$ and $K$, we arrive at three cases where replacement reduces the sequence length: 1) $N \geq 4$, $K \geq 2$, 2) $N=3$, $K \geq 3$, 3) $N=2$, $K \geq 4$.

\noindent\textbf{Subsequence Discovery} (Algorithm \ref{alg:alg1}): To find candidate subsequences for replacement, we search $T$ for subsequences of length two or more up to some fixed maximum subsequence length $L_{max}$. For each candidate subsequence $T_{sub}$, we only count the total number of times $T_{sub}$ occurs without overlapping with itself and save the starting index position of each occurrence. If the count $K$ and length $N$ satisfy Equation \ref{eq:swap_condition}, we add $T_{sub}$ to a list $S$ of candidate subsequences that can be swapped. This approach is similar to the $f$-gram discovery algorithm from \citet{yu2025scaling} and, as discussed in their paper, resembles continued training of a BPE tokenizer.

\noindent\textbf{Subsequence Swapping} (Algorithm \ref{alg:alg2}): We iteratively swap out each subsequence with a unique meta-token $m$ sampled uniformly from the list $M$ of available meta-tokens. Since it is possible for swappable subsequences to overlap with one another, before making a swap we verify that all positions for the swap are still original tokens from $T$ and that Equation \ref{eq:swap_condition} still holds for the given subsequence $T_{sub}$.

\noindent\textbf{Dictionary Construction} (Algorithm \ref{alg:alg3}): To indicate the dictionary in the token sequence format, we concatenate $($meta-token$,$ subsequence$)$ pairs within the XML tags $<$Dict$>$ and $<$/Dict$>$. We then prepend this dictionary to the compressed token sequence.

\subsection{Time Complexity}\label{sec:time_complexity_analysis}

The overall time complexity of the compression process is $O(|T|\log |T|)$, making it scalable for long sequences (See Appendix \ref{sec:runtime_exps} for empirical verification). We discuss sub-step complexity below:

\textbf{1)} \textit{Subsequence Discovery}: Collecting all unique subsequences for a given $T$ is bounded by $|T|\cdot L_{max}$, and filtering these subsequences is bound by the same value. Subsequences are already ordered from longest to shortest since we collect them in that order, therefore this step is $O(|T|)$.

\textbf{2)} \textit{Subsequence Swapping}: To collect the swaps we want to make, we iterate through the set of subsequences that can be swapped. Both the number of unique subsequences and unique starting positions are bounded by $|T|\cdot L_{max}$,
making the swap collection step $O(|T|)$. Applying the swaps is done with one pass through $T$, making this step bounded by $|T|$. The most expensive part of Algorithm \ref{alg:alg2} is sorting the swaps by starting index, making this step $O(|T|\log |T|)$.

\textbf{3)} \textit{Dictionary Construction}: We build the token sequence via one pass through the dictionary; $O(|T|)$.

\section{Training and Inference with Compression}

\noindent\textbf{Model Architecture Changes}: Given a desired number of meta-tokens $|M|$, we must augment the existing LLM vocabulary with $|M|$ new tokens. For a transformer-based LLM with input dimension $d$, this results in the addition of $|M|\cdot d$ new parameters for LLMs using weight tying \cite{press-wolf-2017-using} and $2|M|\cdot d$ for LLMs with separate vocabulary and language modeling head matrices. 

\noindent\textbf{Supervised Fine-Tuning}: All tasks explored in this paper can be formatted as \texttt{(prompt, answer)} tuples. During fine-tuning, we compress the \texttt{prompt} but leave the \texttt{answer} uncompressed, and we only compute the cross-entropy loss over the answer tokens. This forces the model to retain its original output distribution by only generating the answer using tokens from its original vocabulary.\footnote{We discuss the negative effects of computing loss over compressed sequences in Section \ref{sec:future_work}.}

\noindent\textbf{Inference}: At inference time, we compress the \texttt{prompt} and generate until the \texttt{<|endoftext|>} token is produced. For all the tasks explored here, prompt length is bounded by 8k tokens. Based on runtimes from Figure \ref{fig:runtime}, compression introduces at most $\sim$100ms of pre-processing latency.

\section{Experiments}\label{sec:experiments}

We evaluate \texttt{LTSC} and the lossy compression method LLMLingua2 \cite{pan2024llmlingua} on two tasks that require a lossless input representation (lossless tasks). The results highlight two key properties:

\noindent\textbf{1)} \textit{Collapse of Lossy Representations}: For lossless tasks, lossy compression results in poor performance due to inevitable information loss.

\noindent\textbf{2)} \textit{Strong Performance of }\texttt{LTSC}: Our proposed compression technique performs similarly to a vanilla model that does not compress the input on all lossless tasks due to preservation of all information from the original prompt.

\noindent\textbf{Tasks}: We evaluate on two lossless tasks: \textbf{1)} Tree Structure Understanding, \textbf{2)} Repository-level Code Completion \cite{liu2023repobench}. We show that for both tasks, dropping even a small number of tokens from the input leads to significant performance degradation.

\noindent\textbf{Baseline}: We use LLMLingua2 \cite{pan2024llmlingua} as our prompt compression baseline. LLMLingua2 is currently SOTA for lossy compression and works by using an encoder model to classify whether each token in a given sequence is kept or dropped according to a desired compression rate. The encoder is trained via knowledge distillation from compressed prompts generated by GPT-4. LLMLingua2 claims to be task-agnostic and work well with any LLM, but it shows information loss when evaluated on our lossless tasks.

\subsection{Tree Structure Understanding}\label{sec:tree_structure_understanding}

\begin{wrapfigure}{r}{0.52\textwidth}
    \centering
    \includegraphics[width=0.51\textwidth]{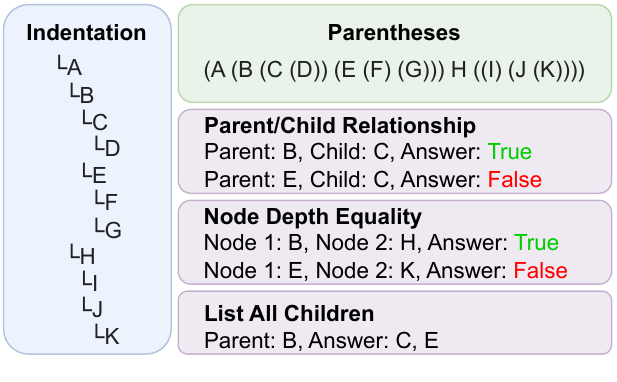}
    \caption{Tree Structure Understanding: A sample tree is shown in both Indentation and Parentheses formats along with examples of each of the tree understanding tasks.}
    \label{fig:lossless_comp_tasks}
\end{wrapfigure}

Trees are a fundamental data structure and can be represented in multiple ways as a text sequence (see Figure \ref{fig:lossless_comp_tasks}). We will demonstrate that accurate understanding of the relationship between nodes requires
perfect preservation of all information in the sequence, because every character contributes to the global structure of the tree. In this experiment, we represent trees using the \textit{Hierarchical Indentation} (Indentation) format or \textit{Parentheses Notation} (Parentheses) format. For Indentation, each node is on a line of its own and node depth is represented by the number of spaces. For Parentheses, tree structure is represented via nested sets of parentheses. Removal of any characters in either representation can render a different tree, making it challenging to understand the tree structure when the textual representation is compressed in a lossy fashion. We show this empirically by devising three toy tasks that probe whether an LLM can understand the structure of a given tree.

\noindent\textbf{1)} \textit{Parent/Child Relationship}: Given two nodes $A$ and $B$, determine whether $A$ is the parent of $B$. This is a binary classification task that returns true if $B$ is a child of $A$ and false otherwise.

\noindent\textbf{2)} \textit{Node Depth Equality}: Given two nodes $C$ and $D$, determine whether the nodes are at the same depth in the tree. This is a binary classification task that returns true if $C$ and $D$ have the same depth and false otherwise.

\noindent\textbf{3)} \textit{List All Children}: Given a parent node $E$, list all children of the node. Since the children can be generated in different orders, correctness is determined by comparing the set of predicted nodes to the ground truth set of children.

\begin{table*}[t!]
    \footnotesize
    \centering
    \begin{tabular}{c c c c c c c }
    \Xhline{3\arrayrulewidth}
    \multicolumn{4}{c}{Experiment Settings} & \multicolumn{3}{c}{Tree Understanding} \\
    \cmidrule{5-7}
    \multicolumn{1}{c}{Comp.} & \multicolumn{1}{c}{Comp.} & \multicolumn{1}{c}{Agv. Len.} & \multicolumn{1}{c}{Tree} & \multicolumn{1}{c}{Parent/Child} & \multicolumn{1}{c}{Node Depth} & \multicolumn{1}{c}{List All} \\
    \multicolumn{1}{c}{Seq.} & \multicolumn{1}{c}{Method} & \multicolumn{1}{c}{Red. (\%) $\uparrow$} & \multicolumn{1}{c}{Repr.} & \multicolumn{1}{c}{Relationship $\uparrow$} & \multicolumn{1}{c}{Equality $\uparrow$} & \multicolumn{1}{c}{Children $\uparrow$} \\
    \cmidrule{5-7}
    $\greencheck$ & LTSC & 27.1 & Indentation & 99.68 & 99.67 & 98.32 \\
    $\greencheck$ & LTSC & 21.4 & Parentheses & 99.83 & 99.91 & 99.09 \\
    $\redxmark$ & None & 0.0 & Indentation & 99.97 & 100.0 & 99.82 \\
    $\redxmark$ & None & 0.0 & Parentheses & 99.95 & 99.99 & 99.87 \\
    $\greencheck$ & LLMLingua2 & 27.0 & Indentation & 82.17 & 72.30 & 36.58 \\
    $\greencheck$ & LLMLingua2 & 17.6 & Indentation & 87.45 & 77.71 & 36.63 \\
    $\greencheck$ & LLMLingua2 & 13.9 & Indentation & 89.75 & 79.70 & 36.62 \\
    $\greencheck$ & LLMLingua2 & 26.6 & Parentheses & 80.77 & 54.25 & 36.99 \\    
    $\greencheck$ & LLMLingua2 & 12.1 & Parentheses & 87.16 & 71.61 & 34.71 \\    
    $\greencheck$ & LLMLingua2 & 5.4 & Parentheses & 90.65 & 81.77 & 44.25 \\
    N/A & Random Guess & N/A & N/A & 50.00 & 50.00 & N/A \\
    
    \Xhline{3\arrayrulewidth}
    \end{tabular}
    \caption{Tree structure understanding results with accuracy given in \%. These results are obtained by fully fine-tuning \texttt{Qwen2.5-Coder-0.5B}.}
    \label{tab:tree_results}
\end{table*}

\noindent\textbf{Data Generation}: We generate random trees with a maximum depth of four and a maximum of 150 nodes, where each node has between three and five children. The node values are sampled without replacement from the set of two-character capital letter sequences (676 possible values). For further details about data generation, refer to Appendix \ref{sec:data_gen_tree}. For Tree Structure Understanding, the prompt consists of the tree in either Indentation or Parentheses format followed by the formatted representation of each task indicated in the examples in Figure \ref{fig:lossless_comp_tasks}.

\noindent\textbf{Model Training}: We fine-tune \texttt{Qwen2.5-Coder-0.5B} for one epoch using a batch size of 128, 500 meta-tokens and maximum subsequence length $L_{max}$ of six. We leave half of the data in its original (uncompressed) format and compress the prompt according to \texttt{LTSC} for the other half. This training recipe allows us to test multiple settings at inference time with one model.

\noindent\textbf{Experiment Settings}: We evaluate three setups: \textbf{1)} \texttt{LTSC}, \textbf{2)} LLMLingua2 compression, \textbf{3)} No compression, i.e. vanilla model. For LLMLingua2, we apply compression to the tree representation part of the prompt only with compression rates of 0.8, 0.9, and 0.95 using the official implementation.\footnote{\url{https://github.com/microsoft/LLMLingua}} Due to tokenizer differences between LLMLingua2's encoder\footnote{\texttt{microsoft/llmlingua-2-xlm-roberta-large-meetingbank}} and \texttt{Qwen2.5-Coder-0.5B}, the compression rates for our fine-tuned model are slightly different. We report the compression values based on \texttt{Qwen2.5-Coder-0.5B}'s tokenizer in Table \ref{tab:tree_results}.

\noindent\textbf{Results/Discussion}: Results for tree understanding are in Table \ref{tab:tree_results}. We highlight two key observations:

\noindent\textbf{1)} \textit{Almost perfect performance from} \texttt{LTSC}: For both the Indentation and Parentheses format, our compression approach achieves near perfect performance. The vanilla baseline (uncompressed input) only slightly outperforms \texttt{LTSC}, with the largest performance gap being less than 1\%. This result empirically validates that although the sequence length is reduced with \texttt{LTSC} by up to 27\%, no information is lost.

\noindent\textbf{2)} \textit{Poor performance for lossy compression}: For all three tree understanding tasks, LLMLingua2 experiences significant performance degradation relative to \texttt{LTSC} even when its compression amount is much smaller. This result verifies that the Tree Structure Understanding tasks are lossless, because the removal of a small fraction of tokens from the tree representation makes it difficult to perform each task.

\noindent\textbf{Trends for Lossy Compression}: Performance by LLMLingua2 varies widely across the three tasks, indicating that some tree understanding tasks are more forgiving than others when structural information is lost. We compare the three tasks below:

\noindent\textbf{1)} \textit{Best Performance}: LLMLingua2 achieves the best performance on the \textit{Parent/Child Relationship} task. This is likely due to the fact that a node's proximity to a reference parent node is a good feature for predicting a parental relationship in both Indentation and Parentheses formats.

\noindent\textbf{2)} \textit{Worst Performance}: Regardless of how small the amount of compression is for LLMLingua2, performance on the \textit{List All Children} task is always poor. Given the strict requirement that all children must be produced for a response to be counted as correct, dropping only a few tokens will frequently either corrupt the tree structure or alter at least one of the node values.

\noindent\textbf{3)} \textit{Indentation vs. Parentheses}: LLMLingua2 achives much better performance on the Node Depth Equality task when using the Indentation vs. Parentheses format (72.30\% vs. 54.25\%) for the most extreme LLMLingua2 compression setting. The only feature needed to determine node depth for Indentation is the number of spaces next to a given node's value, whereas the nested structure of the Parentheses format makes it much easier to corrupt node depth information.

\begin{table*}[t!]
    \scriptsize
    \centering
    \begin{tabular}{c c c c c c c c c c c }
    \Xhline{3\arrayrulewidth}
    \multicolumn{3}{c}{Experiment Settings} & \multicolumn{8}{c}{RepoBench} \\
    \cmidrule{4-11}
    \multicolumn{1}{c}{} & \multicolumn{1}{c}{Comp.} & \multicolumn{1}{c}{Agv. Len.} & \multicolumn{2}{c}{XF-F} & \multicolumn{2}{c}{XF-R} & \multicolumn{2}{c}{IF} & \multicolumn{2}{c}{Average}\\
    \multicolumn{1}{c}{Model} & \multicolumn{1}{c}{Method} & \multicolumn{1}{c}{Red. (\%) $\uparrow$} & \multicolumn{1}{c}{EM $\uparrow$} & \multicolumn{1}{c}{ES $\uparrow$} & \multicolumn{1}{c}{EM $\uparrow$} & \multicolumn{1}{c}{ES $\uparrow$} & \multicolumn{1}{c}{EM $\uparrow$} & \multicolumn{1}{c}{ES $\uparrow$} & \multicolumn{1}{c}{EM $\uparrow$} & \multicolumn{1}{c}{ES $\uparrow$}   \\
    
    \cmidrule{4-11}
     & LTSC & 17.9 & 18.9 & 0.67 & 28.5 & 0.72 & 25.9 & 0.67 & 24.5 & 0.69 \\
    \texttt{StarCoder2-3B} & None & 0 & 24.6 & 0.72 & 37.5 & 0.77 & 34.4 & 0.71 & 32.2 & 0.73 \\
     & LLMLingua2 & 8.7 & 0.5 & 0.59 & 0.5 & 0.63 & 0.3 & 0.55 & 0.4 & 0.59 \\
    \Xhline{1\arrayrulewidth}
     & LTSC & 17.9 & 22.6 & 0.70 & 30.5 & 0.74 & 30.5 & 0.69 & 27.9 & 0.71 \\
    \texttt{StarCoder2-7B} & None & 0 & 27.8 & 0.74 & 38.8 & 0.78 & 38.9 & 0.73 & 35.2 & 0.75 \\
     & LLMLingua2 & 8.7 & 0.0 & 0.63 & 0.0 & 0.64 & 0.2 & 0.56 & 0.0 & 0.61 \\
    \Xhline{1\arrayrulewidth}
     & LTSC & 17.9 & 27.5 & 0.72 & 36.5 & 0.76 & 34.8 & 0.72 & 32.9 & 0.73 \\
    \texttt{StarCoder2-15B} & None & 0 & 30.8 & 0.77 & 39.5 & 0.79 & 40.2 & 0.73 & 36.9 & 0.76 \\
     & LLMLingua2 & 8.7 & 0.5 & 0.65 & 0.3 & 0.66 & 0.2 & 0.58 & 0.3 & 0.63 \\
    \Xhline{1\arrayrulewidth}
     & LTSC & 15.7 & 8.5 & 0.59 & 12.3 & 0.62 & 12.6 & 0.56 & 11.2 & 0.59 \\
    \texttt{Qwen2.5-Coder-0.5B} & None & 0 & 15.7 & 0.66 & 29.0 & 0.71 & 29.0 & 0.66 & 24.6 & 0.68 \\
     & LLMLingua2 & 15.2 & 0.0 & 0.50 & 0.2 & 0.55 & 0.0 & 0.46 & 0.1 & 0.51 \\
    \Xhline{1\arrayrulewidth}
      & LTSC & 15.7 & 17.9 & 0.67 & 29.8 & 0.72 & 28.2 & 0.68 & 25.3 & 0.69 \\
    \texttt{Qwen2.5-Coder-1.5B}  & None & 0 & 22.9 & 0.70 & 34.2 & 0.76 & 33.8 & 0.70 & 30.3 & 0.72 \\
     & LLMLingua2 & 15.2 & 0.2 & 0.59 & 0.2 & 0.60 & 0.3 & 0.52 & 0.2 & 0.57 \\
    \Xhline{1\arrayrulewidth}
      & LTSC & 15.7 & 18.9 & 0.69 & 28.3 & 0.72 & 26.6 & 0.66 & 24.6 & 0.69 \\
    \texttt{Qwen2.5-Coder-3B}  & None & 0 & 28.4 & 0.74 & 35.7 & 0.76 & 37.9 & 0.73 & 34.0 & 0.74 \\
     & LLMLingua2 & 15.2 & 0.3 & 0.62 & 0.3 & 0.61 & 0.5 & 0.55 & 0.4 & 0.59 \\   
    
    \Xhline{3\arrayrulewidth}
    \end{tabular}
    \caption{RepoBench results for the 2k/4k/8k bins (combined). Exact Match (EM) accuracy is given in \%, and Edit Similarity (ES) ranges from 0-1. EM requires the generated code to exactly match the gold reference to be counted correct in the accuracy calculation. ES is the average string similarity between the generated and gold code.}
    \label{tab:repobench_results}
\end{table*}

\subsection{Repository-level Code Completion}\label{sec:repo_level_code_completion}

One common application of LLMs is to act as a coding assistant by predicting the next line of code for a user. This task is called Repository-level Code Completion, and we explore an LLM's ability to complete this task using \texttt{LTSC} across a variety of model scales. We use the RepoBench dataset \cite{liu2023repobench} for training and evaluation, where we limit data to the 2k, 4k, and 8k context length bins only due to a context length limit of 10,240 tokens during training. RepoBench contains Python and Java repositories where examples have been curated into three settings for next-line prediction: \textbf{1)} Cross-File-First (XF-F), \textbf{2)} Cross-File-Random (XF-R), and \textbf{3)} In-File (IF). We split RepoBench by repository and allocate 80\% of repos for training, 10\% for validation, and 10\% for testing. The quality of the predicted lines are measured using Exact Match (EM) and Edit Similarity (ES).

\noindent\textbf{Fine-Tuning}: We find empirically that it takes several hundred weight updates for an LLM to learn the input transformation induced by \texttt{LTSC}. To directly teach the model to use the dictionary at the beginning of the prompt, we use Python and Java programs from an instruction-tuning dataset called CommitPack \cite{muennighoff2024octopackinstructiontuningcode}. The data consists of GitHub Pull Requests (PR), where the file prior to the PR is the \texttt{input}, the \texttt{instruction} consists of the PR comments, and the \texttt{answer} is the file after the PR. For this task, the prompt is constructed by concatenating \texttt{instruction} and \texttt{input}. Since most of the file is left unchanged in each PR, the LLM must learn to use the dictionary pairs as well as reason about necessary file changes to be able to produce the \texttt{answer} correctly.

\noindent\textbf{Experiment Settings}: We fine-tune all models using LoRA \cite{hu2021lora} for 630 steps on CommitPack, followed by 80 steps on RepoBench where the max seq. length is 10,240. We find the optimal RepoBench validation loss after 30 steps, and choose this checkpoint for every fine-tuned model. We train two separate versions of each LLM,\footnote{Training separate models rules out the possibility that \texttt{LTSC} negatively impacts the performance of the vanilla model. This was unnecessary for the previous task since the vanilla model achieves almost perfect performance.} one with \texttt{LTSC} and one vanilla model that does not use compression. We use 500 meta-tokens, set $L_{max}=6$, and finetune \texttt{Qwen2.5-Coder-(0.5,1.5,3)B} and \texttt{StarCoder2-(3,7,15)B}. To evaluate LLMLingua2 \cite{pan2024llmlingua}, we compress the prompt at inference time and generate responses using the vanilla model.

\noindent\textbf{Results/Discussion}: Repo-level code completion results are in Table \ref{tab:repobench_results}. We note two key points:

\noindent\textbf{1)} \texttt{LTSC} \textit{outperforms LLMLingua2}: Even with a larger amount of compression, \texttt{LTSC} is able to outperform LLMLingua2 on code completion. As expected, LLMLingua2 is unable to achieve even 1\% on Exact Match due to the loss of important syntax/structural information. For Edit Similarity, LLMLingua2 achieves reasonable performance but still falls behind \texttt{LTSC} by $\sim$0.1 on average for most models.

\begin{wrapfigure}{r}{0.52\textwidth}
    \centering
    \includegraphics[width=0.51\textwidth]{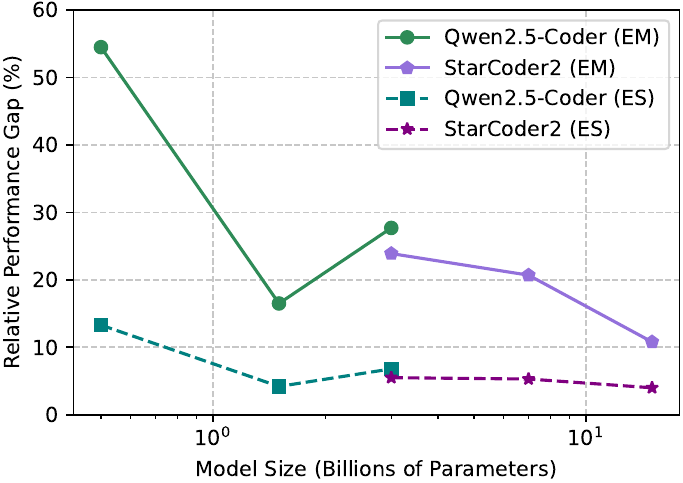}
    \caption{Relative performance gap between \texttt{LTSC} and vanilla model (no compression) on repository-level code completion for a variety of model sizes.}
    \label{fig:performance_gap_scale}
\end{wrapfigure}

\noindent\textbf{2)} \textit{Scaling Trend}: Although \texttt{LTSC} does not achieve the same performance as the vanilla model, the relative performance gap shrinks significantly as the size of the LLM increases (See Figure \ref{fig:performance_gap_scale}). This demonstrates that larger models are able to understand \texttt{LTSC}'s input transformation better than smaller models when fine-tuning for a relatively small number of steps ($<$1k).

\section{Discussion}

The results from both tasks provide strong evidence for the need for lossless compression methods. In every case, we see that the lossy baseline experiences a large degradation in performance even for small amounts of compression compared to the settings from previous studies \cite{pan2024llmlingua, jiang2023llmlingua}. Each task requires preservation of structural and syntactical information that is corrupted when LLMLingua2 drops a relatively small number of tokens. We also demonstrate that transformer-based LLMs are capable of understanding the input transformation induced by \texttt{LTSC} and can perform tasks at a similar level
to vanilla models not using compression. We find the relative performance gap between \texttt{LTSC} and vanilla models becomes significantly smaller with scale, suggesting that LLMs with hundreds of billions of parameters may not experience any performance gap.

\noindent\textbf{Why does compression work?} Unlike recurrent models like RNNs \cite{schuster1997bidirectional} or LSTMs \cite{hochreiter1997long} where sequential order is inherent to the model, Transformers operate on sets \cite{vaswani2017attention}, so positional information is injected artificially through position embeddings. For our approach, the compressed sequence contains the same underlying information as the original sequence, but has been rearranged to reduce the sequence length. Given the simplicity of the transformation and the incredible abilities of LLMs to solve challenging tasks, it is expected that LLMs should understand this simple input transformation.

\section{Future Work}\label{sec:future_work}

\noindent\textbf{Model Scale}: Our experimental results have demonstrated that \texttt{LTSC}'s input transformation can be learned well with a relatively small amount of compute. For the largest model, \texttt{StarCoder2-15B}, by fine-tuning with LoRA for less than 1k steps, we only observe a relative 10.9\% Exact Match (EM) gap between \texttt{LTSC} and the uncompressed prompt on the challenging code completion task. Additionally, the performance gap is halved for StarCoder2 as the model scale increases by less than one order of magnitude from three to fifteen billion parameters (See Figure \ref{fig:performance_gap_scale}). Given the significant impact scale has on language modeling performance \cite{kaplan2020scaling}, we hypothesize that applying our approach to larger models and fine-tuning longer will likely remove the performance gap entirely.

\noindent\textbf{Generating Meta-tokens}: In addition to reducing the input sequence length, using meta-tokens in the output could further speed up generation and significantly reduce cost, since output tokens are currently 4-5 times more expensive than input tokens from the top AI companies. Our initial experiments in this area were not promising, which we hypothesize is due to two reasons: \textbf{1)} Predicting meta-tokens instead of original vocabulary tokens causes a significant distribution shift. \textbf{2)} Probability mass for a given character sequence becomes distributed over multiple tokens since meta-tokens make it possible to represent the underlying character sequence in multiple ways. Future large-scale training with \texttt{LTSC} could overcome these issues, but this is left to future work.

\section{Conclusions}

In this paper, we introduced the first lossless compression algorithm for token sequences and demonstrated a need for lossless input representations. While previous work on lossy compression achieves solid performance with high compression ratios, those same methods suffer when applied to the information-dense tasks we explore in this paper. In contrast, we found little performance gap between our proposed compression approach and a no-compression baseline for the largest model, and found that this gap shrinks significantly with model scale.

The work explored in this paper marks a fundamental shift in the way we think about input for transformer-based LLMs. While left-to-right structure is required for simpler recurrent models, we have demonstrated empirically that this is not the case for transformers because attention acts as a set operator. This freedom to rearrange the input allows us to capitalize on a denser representation of the same token sequence, and we have shown that transformer-based LLMs can use this compact representation effectively to solve downstream tasks.

\bibliographystyle{plainnat}
\bibliography{bibliography}

\newpage
\appendix
\onecolumn

\begin{algorithm}[t!]
\small
\caption{Subsequence Discovery}\label{alg:alg1}
\begin{algorithmic}
    \State{Token sequence $T=(t_1, t_2, ..., t_{|T|})$, maximum subsequence length $L_{max}$, non-overlapping subsequence starting indices $I=(i_1, i_2, ..., i_K)$. Denote subsequences via slicing: $T_{sub[i:j]}=(t_i, t_{i+1}, ..., t_{j-1})$.}    
    \Function{SubsequenceDiscovery}{$T$, $L_{max}$}
    
    \State $S \leftarrow$ \text{Empty list (holds subsequences and their start indices)}
    
    \For{$N \in (L_{max}, L_{max} -1, ..., 2)$}
        \State $U_N \leftarrow$ \text{Set of all unique subsequences of length $N$ in $T$}
        
        \For{$T_{sub} \in U_N$}

            \State $I \leftarrow$ \text{List of starting indices for each non-overlapping occurrence of } $T_{sub}$ in $T$

            \State $S \leftarrow S + (T_{sub}, I)$
            
        \EndFor
    \EndFor

    \State{Filter $S$ by Equation \ref{eq:swap_condition}: $|I| \cdot |T_{sub}| > |I| + |T_{sub}| + 1$}
    
    \State \Return $S$
    
    \EndFunction
\end{algorithmic}
\end{algorithm}

\begin{algorithm}[t!]
\small
\caption{Sequence Compression}\label{alg:alg2}
\begin{algorithmic}
    \State{Token sequence $T=(t_1, t_2, ..., t_{|T|})$, list $S$ of subsequences $T_{sub}$ and their starting indices $I$, meta-tokens $M=(m_1, m_2, ..., m_{|M|})$, dictionary of compression pairs $D$. }    
    \Function{CompressTokenSequence}{$T$, $S$}

    \State $D \leftarrow$ \text{Dict[key is meta-token $m$, value is subsequence $T_{sub}$]}

    \State $I_{swap} \leftarrow$ \text{Empty list (holds all index positions that have been swapped)}

    \State $B \leftarrow$ \text{Empty list (holds all the swaps that will be performed on the original token sequence)}

    \For{$(T_{sub}, I) \in S$} \Comment{subsequence $T_{sub}$, starting indices $I$}

    \State $I \leftarrow I \setminus I_{swap}$ 
    \Comment{Remove invalid swap positions by checking that no index within [$I$, $I+|T_{sub}|-1$] belongs to $I_{swap}$.}

    \If{$|I| \cdot |T_{sub}| > |I| + |T_{sub}| + 1$}

    \If{$M$ not empty}

    \State Sample $m$ from $M$ without replacement.

    \For{$i \in I$}

        \State $B \leftarrow B \cup \{(i, T_{sub})\}$ 
\Comment{Add current start index and subsequence pair to list of swaps that will be performed.}

        \State $I_{swap} \leftarrow I_{swap} \cup \{i, ..., i+|T_{sub}|-1\}$ 
\Comment{Add all indices from the swap span to the list of used indices.}

    \EndFor

    \State $D[m] \leftarrow T_{sub}$ 
\Comment{Add meta-token/subsequence mapping pair to the dictionary.}

    \EndIf

    \EndIf

    \EndFor

    \State \text{Sort} $B$ \text{from smallest to largest start index.}

    \State \text{Apply swaps in} $B$ \text{to} $T$ \text{(left-to-right). Keep track of the position shift as $T$ is modified.}
    \State \Return $T, D$
    \EndFunction
\end{algorithmic}
\end{algorithm}

\begin{algorithm}[]
\small
\caption{Input Sequence Construction}\label{alg:alg3}
\begin{algorithmic}
    \State{Compressed token sequence $T_{comp}=(t_1, t_2, ..., t_{|T_{comp}|})$, subsequences $T_{sub}$, meta-tokens $M=(m_1, m_2, ..., m_{|M|})$, dictionary of meta-token/subsequence compression pairs $D$. }    
    \Function{ConstructInput}{$T_{comp}$, $D$}

    \State $T_{dict} \leftarrow$ Empty list (holds token representation of the dictionary)

    \For{($m$, $T_{sub}$) in $D$}

    \State $T_{dict} \gets T_{dict} + m + T_{sub}$

    \EndFor

    \State $T_{comp} \gets$ \text{[$<$Dict$>$]} $+ T_{dict} +$ \text{[$<$/Dict$>$]} $+T_{comp}$

    \State \Return $T_{comp}$
    \EndFunction
\end{algorithmic}
\end{algorithm}

\section{Related Work}
\label{related_work_appendix}

\subsection{Hard Prompt Compression}

The most prominent work on hard prompt compression comes from the LLMLingua family of methods \cite{jiang2023longllmlingua, jiang2023llmlingua, pan2024llmlingua}. The first version of LLMLingua \cite{jiang2023llmlingua} computes perplexity for each token using a small model and drops those tokens with perplexity below a given threshold. This approach gives an empirical estimation of the tokens that are least informative for the LLM. LLMLingua is improved upon by LLMLingua2 \cite{pan2024llmlingua}, which uses knowledge distillation from a much larger model. Given prompts that have been compressed by GPT-4, a small, bi-directional encoder is trained to classify tokens that should be discarded. LLMLingua2 is currently State-Of-The-Art (SOTA) for lossy compression and is used as the baseline in this paper.

The second form of hard prompt compression involves generating a prompt summary in natural language, which results in significant latency overhead compared to the first form. Nano-capsulator \cite{chuang-etal-2024-learning} optimizes the summarized prompt to retain utility for the downstream task and achieves superior performance compared to task-agnostic approaches. This performance boost comes at the cost of being unable to transfer to new tasks without further fine-tuning.

\subsection{Soft Prompt Compression}
Both GIST \cite{mu2024learning} and ICAE \cite{ge2023context} introduce insightful methods by which prompts can be compressed when we allow the underlying LLM to be further fine-tuned. GIST \cite{mu2024learning} bridges the gap between fine-tuning and prompt tuning by producing compressed versions of task instructions in a zero-shot setting. During fine-tuning, GIST alters the attention mask such that during generation of an answer, the LLM can only attend to a small set of gist (summarized) tokens whose length is reduced from the original instruction prompt. ICAE \cite{ge2023context} develops a similar approach, where an encoder produces memory slot tokens that are conditioned on by a frozen decoder LLM. The encoder learns to compress useful information from a long prompt into the limited number of memory slot tokens via auto-encoding, where the frozen decoder conditions on the memory tokens and attempts to reconstruct the original prompt.

\begin{figure}[t!]
    \centering
    \includegraphics[width=10cm]{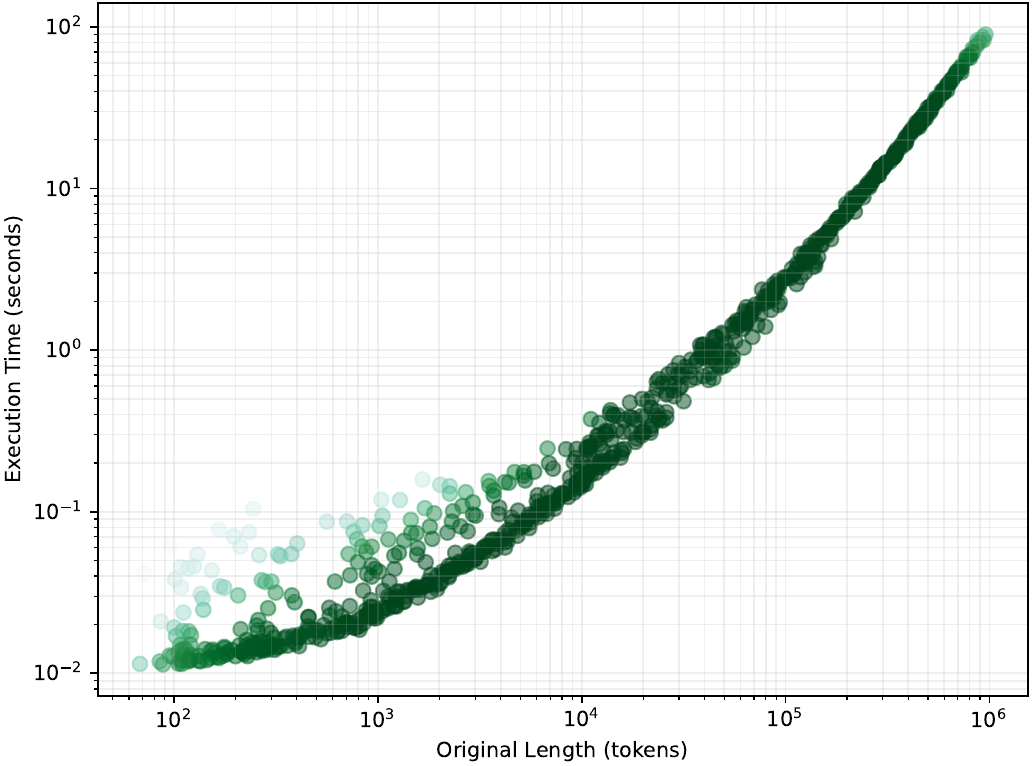}
    \caption{Runtime of compression process for Wikipedia data tokenized by \texttt{Qwen2.5-Coder}.}
    \label{fig:runtime}
\end{figure}

\section{Runtime Experiments}\label{sec:runtime_exps}

The time complexity of $O(|T|\log |T|)$ for the compression process is empirically verified in Figure \ref{fig:runtime}. We concatenate articles from Wikipedia to construct natural language sequences of various lengths, up to 1M tokens. As the sequence length grows four orders of magnitude, so does the execution time. We also observe that the variance in execution time becomes smaller as the sequence length becomes large.

\section{Data Generation}

\subsection{Tree Structure Understanding}\label{sec:data_gen_tree}

\noindent\textbf{Parent/Child Relationship}: For this task, we generate positive data pairs by randomly selecting a non-terminal node and then randomly selecting one of its children. We generate negative data pairs by sampling a node at random and then sampling from the set of all other nodes that are neither its parent nor children.

\noindent\textbf{Node Depth Equality}: We generate positive pairs by randomly sampling a depth between two and four and then randomly sampling two nodes from that depth. To create negative pairs, we sample nodes from two different depths.

\noindent\textbf{List All Children}: Given the set of all non-terminal nodes, we randomly sample one node and then sort the children in alphabetical order. We do not require that the nodes be generated in alphabetical order at inference time but provide this structure during training to provide a deterministic order to the set.

We generate a training dataset of 3M examples and a test dataset of 50k examples (10k/10k positive/negative pairs each for Parent/Child Relationship and Node Depth Equality, and 10k examples for List All Children).

\section{Analysis of the Dictionary}

\begin{figure}
    \centering
    \includegraphics[width=10cm]{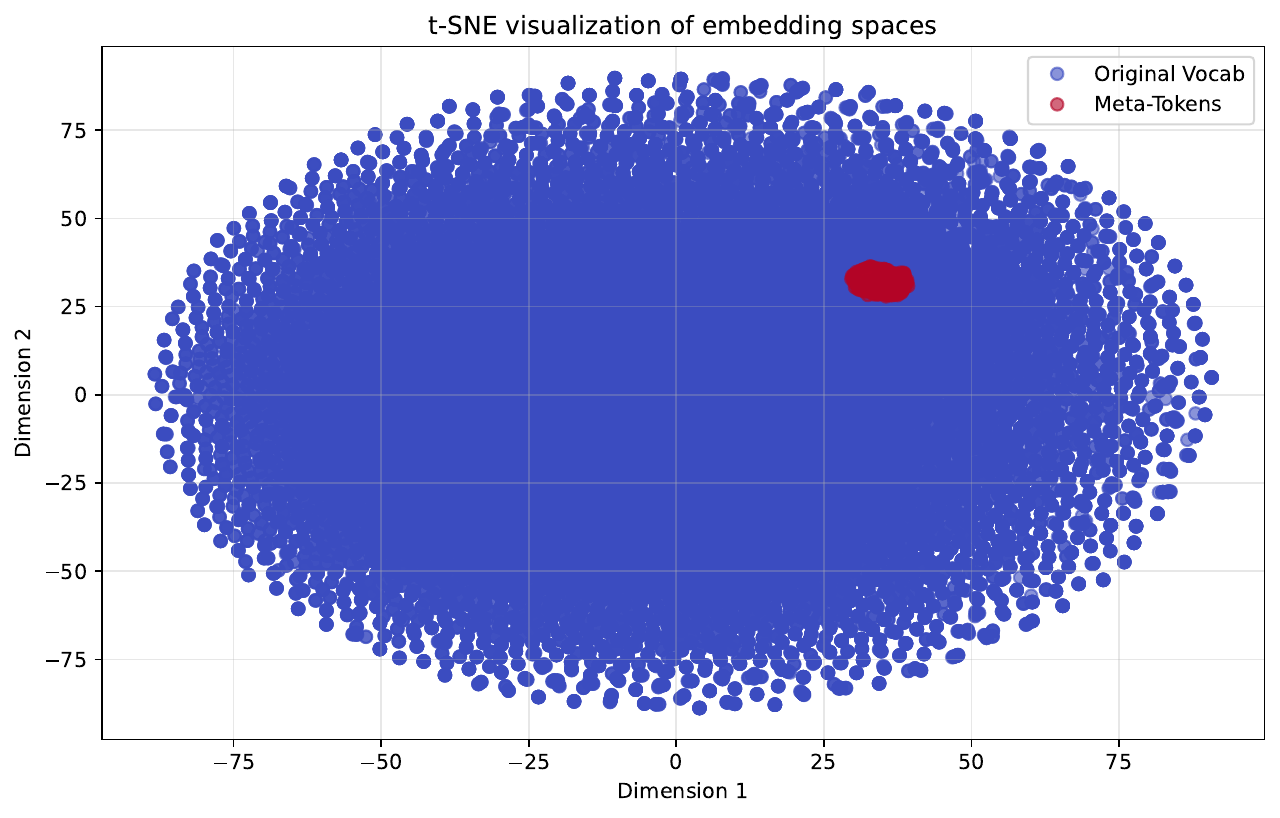}
    \caption{t-SNE plot of meta-tokens and original vocabulary tokens for the StarCoder2-7B model fine-tuned on code completion.}
    \label{fig:t_sne_dictionary}
\end{figure}

\noindent\textbf{Meta-token Embeddings}: We perform a t-SNE analysis \cite{van2008visualizing} on the embeddings of the meta-tokens vs. the original vocabulary tokens and show the results in Figure \ref{fig:t_sne_dictionary}. As expected, the meta-tokens cluster tightly together given that they serve identical purposes for the LLM.

\begin{figure}
    \centering
    \includegraphics[width=10cm]{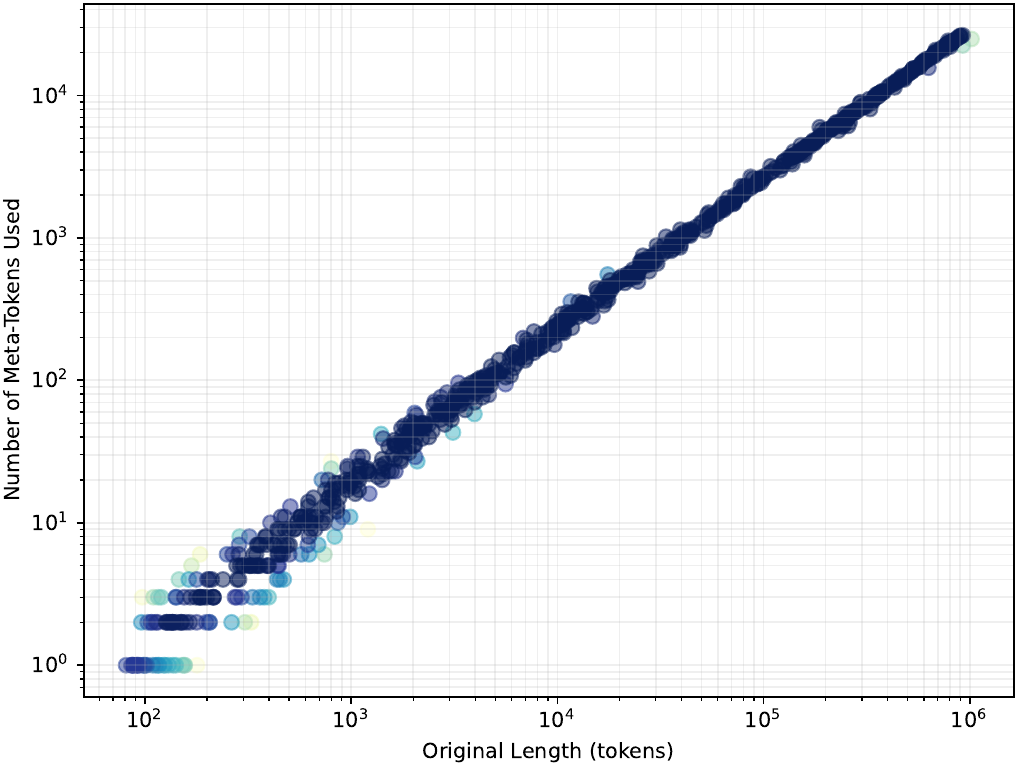}
    \caption{Number of dictionary tokens used by \texttt{LTSC} for natural language token sequences of varying lengths.}
    \label{fig:num_dict_tokens}
\end{figure}

\noindent\textbf{Meta-token Scaling}: We explore the number of meta-tokens that are used when compressing natural language token sequences according to \texttt{LTSC} in Figure \ref{fig:num_dict_tokens}. We see a clear linear trend, that as the token sequence length increases, so does the number of unique meta-tokens used for compression. Given that the maximum sequence length used for our experiments was 8k (code completion), using 500 meta-tokens for the model is a good choice, since all sequences of length 10k use between 200 and 300 meta-tokens. 

\section{Computational Resources}\label{sec:compute_resources}

All experiments are run on one EC2 G6e instance from AWS (8$\times$46GB NVIDIA L40S GPUs, 1.5TB RAM).

\section{Dataset License}\label{sec:data_licenses}

Repobench \cite{liu2023repobench} maintains a Creative Commons license. CommitPack \cite{muennighoff2024octopackinstructiontuningcode} maintains an MIT license.

\end{document}